\documentclass[letterpaper, 10 pt, conference]{ieeeconf}  
\IEEEoverridecommandlockouts      
\usepackage{times}
\usepackage{epsfig}
\usepackage{amsmath}
\usepackage{amssymb}
\usepackage{subcaption}
\usepackage{graphicx}
\usepackage{multirow}
\usepackage{multicol}
\usepackage{makecell}
\usepackage{booktabs}

\title{\LARGE \bf
Dual-Branch Reconstruction Network for Industrial Anomaly Detection with RGB-D Data
}

\author{Chenyang Bi$^{1}$, Yueyang Li$^{1,*}$ and  Haichi Luo$^{2}$ 
\thanks{*Corresponding author.}
\thanks{$^{1}$Jiangsu Provincial Engineering Laboratory of Pattern Recognition and     Computational Intelligence, Jiangnan University, Wuxi, China
    {\tt\small lyueyang@jiangnan.edu.cn}}
\thanks{$^{2}$College of Internet of Things Engineering, Jiangnan University, 1800     Lihu Avenue, Wuxi, Jiangsu, China}
}

\begin{document}

\maketitle

\thispagestyle{empty}

\pagestyle{empty}

\begin{abstract}

Unsupervised anomaly detection methods are at the forefront of industrial anomaly detection efforts and have made notable progress. Previous work primarily used 2D information as input, but multi-modal industrial anomaly detection based on 3D point clouds and RGB images is just beginning to emerge. The regular approach involves utilizing large pre-trained models for feature representation and storing them in memory banks. 
However, the above methods require a longer inference time and higher memory usage, which cannot meet the real-time requirements of the industry. 
To overcome these issues, we propose a lightweight dual-branch reconstruction network(DBRN) based on RGB-D input,  learning the decision boundary between normal and abnormal examples. 
The requirement for alignment between the two modalities is eliminated by using depth maps instead of point cloud input.
Furthermore, we introduce an importance scoring module in the discriminative network to assist in fusing features from these two modalities, thereby obtaining a comprehensive discriminative result.
DBRN achieves  92.8\% AUROC with high inference efficiency on the MVTec 3D-AD dataset without large pre-trained models and memory banks.

\end{abstract}

\section{INTRODUCTION}

\begin{figure}[t]
  \centering
   \includegraphics[width=1\linewidth]{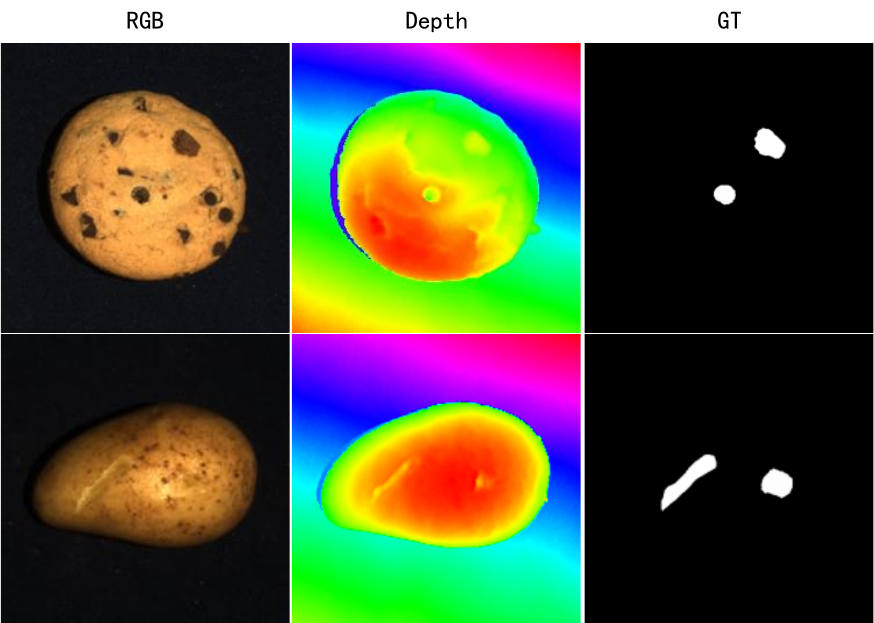}
   \caption{Some examples of defects in RGB images and depth maps from MVTEC 3D-AD.}
   \label{fig:1}
\end{figure}

Industrial surface anomaly detection aims to identify irregularities on the surface of objects, ensuring the quality assurance of industrial products.
In real-world applications, the scarcity of abnormal samples and the diverse range of defect types lead to a preference for using unsupervised methods, where only normal samples are used for training. 
In most scenarios, RGB information is sufficient for locating anomalies, but enhancing it with 3D information can lead to better results.
For example, as shown in Figure~\ref{fig:1}, in the case of the cookie and potato from the MVTEC 3D-AD, the defect is not apparent in the RGB image but becomes visible as a hole in the depth map. Consequently, there is a strong rationale for investigating 3D anomaly detection methods.

The core idea of unsupervised anomaly detection is to identify the differences between abnormal and normal features. Currently, in 2D industrial anomaly detection, there are two main types of methods:~(1) Pre-trained feature extractor-based methods. 
Due to the abundance of positive samples, the extracted features statistically follow a normal distribution. On the one hand, density-based methods~\cite{defard2021padim,kim2021semi,zheng2022focus,jang2023n,bae2022image} utilize statistical features to identify features outside the distribution as anomalies. On the other hand, knowledge distillation-based methods~\cite{bergmann2020uninformed,salehi2021multiresolution,deng2022anomaly,rudolph2023asymmetric,cao2023collaborative} in anomaly detection make judgments based on the disparities in feature representations between teacher network and student network concerning abnormal positions. (2) Reconstruction-based methods. In the reconstruction-based approach~\cite{zavrtanik2021draem,li2020superpixel,pirnay2022inpainting,yan2021learning,liang2023omni}, the features of the original image are first extracted and then used to reconstruct the original image. Since the model is trained with normal samples, it has learned to represent normal patterns accurately. Consequently, when faced with abnormal samples, the model can still attempt to reconstruct them as normal patterns. Discrepancies can be identified by comparing the original images and the reconstructed images, and the defects' location can be determined.

In RGB+3D multimodal approaches, BTF~\cite{horwitz2023back}, CPMF~\cite{cao2023complementary}, and M3DM~\cite{wang2023multimodal} utilize large pre-trained models and memory banks, which result in significantly slow inference speeds and consume substantial memory space. To avoid these issues, we develop a lightweight dual-branch reconstruction network (DBRN) for reconstructing RGB-D images using a self-supervised approach. 
Specifically, DBRN uses the dual-branch reconstruction network to separately reconstruct the RGB image and depth map and use the discriminative network to obtain anomaly localization. We employ masking specific regions or simulating anomalies for data augmentation, restoring the corresponding areas to their original state. 

Combining 2D and 3D modal data to achieve the best results is a common challenge in multimodal data fusion.
In 3D object detection tasks, various fusion methods exist for RGB images and point clouds. Painted PointRCNN~\cite{vora2020pointpainting} uses projection to concatenate the RGB segmentation results with the point cloud. EPNet~\cite{huang2020epnet} uses LI-Fusion layers to fuse Image Features and LiDAR Points. We adopt a reconstruction-based approach using RGB-D data to facilitate feature fusion, using depth maps instead of point clouds. This approach offers two key advantages. 
Firstly, it ensures the natural alignment of the RGB image and depth map at the pixel level, preserving crucial positional information.
Secondly, the depth map being in image format enables us to seamlessly utilize 2D feature extractors and adapt existing 2D methods for processing.
Regarding the fusion of RGB and depth features, we introduce an importance scoring module (ISM) to assess the significance of RGB features and depth features. We utilize a classifier to perform binary classification on RGB features and depth features. The classification results serve as respective importance scores, which are then used to weight the individual features. This process allows us to obtain the fused features. 
Our model achieved a fast inference speed without the use of any large pre-trained models and memory banks.
Experimental results validate the reasonability and effectiveness of our method.

Our contributions are summarized as follows:
\begin{itemize}
\item[$\bullet$] We propose DBRN, a self-supervised dual-branch reconstruction network with high inference speed that utilizes RGB-D information for industrial anomaly detection.
\item[$\bullet$] We propose an anomaly simulation approach for depth maps and simultaneously improve the loss function, resulting in enhanced reconstruction performance.
\item[$\bullet$] We design an importance scoring module (ISM) for RGB and depth features, effectively combining information from both modalities.
\end{itemize}

\section{RELATED WORK}

\subsection{2D Industrial Anomaly Detection}

2D anomaly detection is the mainstream method in industrial anomaly detection, and it has achieved a range of achievements with good performance and outcomes. 2D anomaly detection employs images as input and can currently be categorized into two types of methods.

\textbf{Pre-trained feature extractor-based methods.}
This category of methods utilizes a pre-trained feature extractor to extract features. Features from positive samples objectively conform to a normal distribution, and any features deviating from this distribution are identified as anomalies. The initial approach employed density-based feature statistical methods~\cite{defard2021padim,kim2021semi,zheng2022focus}, capturing the feature distribution of normal samples. The Mahalanobis distance was used to measure feature distribution, calculating the mean and covariance of normal samples. The anomalous image's feature would deviate from this distribution. Recent advancements include the use of normalization flows~\cite{gudovskiy2022cflow,rudolph2022fully,yu2021fastflow}, where a normal feature distribution is explicitly constructed based on a pre-trained network. This method has shown good performance in terms of speed and accuracy. Another approach~\cite{roth2022towards,lee2022cfa} use memory banks to store representative features. During testing, there are significant feature differences when comparing the abnormal features with those normal features stored in the memory bank. Additionally, the teacher-student network approach~\cite{bergmann2020uninformed,salehi2021multiresolution,rudolph2023asymmetric,cao2023collaborative} based on knowledge distillation was also employed. In anomaly detection, knowledge distillation involves training teacher-student networks on normal samples. Due to architectural differences between the teacher and student networks, their capabilities for representing features of anomaly samples also differ. This difference can be used to obtain anomaly scores.

\textbf{Reconstruction-based methods.}
The fundamental assumption of reconstruction methods is the ability to restore anomalous regions. It is achieved by comparing the reconstructed image with the original image, revealing evident differences at the anomalous positions and enabling defect localization. The majority of reconstruction-based methods employ autoencoders(AE)~\cite{bergmann2018improving} or Variational Autoencoders (VAEs)~\cite{liu2020towards} as their reconstruction architecture. AnoGAN~\cite{schlegl2017unsupervised} was the pioneer in introducing GANs into anomaly detection.~\cite{li2020superpixel, yan2021learning, pirnay2022inpainting} mask spectial regions and reconstruct those areas for judgment.
However, during inference, it is necessary to use superpixels to mask the image and then restore them one by one, which poses significant challenges to speed and accuracy.
To enhance the network's capability in distinguishing normal and abnormal features,~\cite{li2021cutpaste,zavrtanik2021draem} employ simulated anomalies, which enable learning decision boundaries for both normal and abnormal samples.
In~\cite{yan2021learning,zavrtanik2021draem,liang2023omni}, a discriminative network is employed for auxiliary judgment. 
The memory bank-based reconstruction approachs~\cite{yang2023memseg,gong2019memorizing} remain effective, wherein normal sample features are stored.
SSPCAB~\cite{ristea2022self} introduced self-supervised prediction architecture blocks that can be universally inserted into reconstruction networks, achieving favorable outcomes. Despite employing various optimization techniques, these reconstruction methods yield inferior results compared to other anomaly detection methods. 

The primary limitations stem from either the networks' restricted feature extraction capacity or anomaly generation strategy issues.
Therefore, we employ a lightweight transformer-based network to enhance the performance of the reconstruction network. We also use anomaly simulation and weighted loss to improve the network's ability to reconstruct abnormal locations. 
Due to the excellent performance of the reconstruction network, we opt for a shallow discriminative network, which is sufficient for defect detection.
Finally, our network achieves a high inference speed without a large model and memory banks.

\subsection{3D Industrial Anomaly Detection}

3D anomaly detection is an extension of 2D anomaly detection. 
In most scenarios, RGB information is sufficient for locating anomalies, but enhancing it with 3D information can lead to better results.
Therefore, it is necessary to integrate the information from both modalities to obtain a comprehensive assessment. When using multiple modalities, we need to address the fusion of information from these two modalities.

Early attempts introduced teacher-student networks~\cite{bergmann2023anomaly} based on 3D point clouds, leveraging additional datasets to construct an expressive teacher network that extracts dense local geometric descriptors. However, it exhibited subpar performance. Subsequent teacher-student network~\cite{rudolph2023asymmetric} solutions prioritize RGB information with depth information as a supplementary multi-modal feature. It involves connecting the depth map to the RGB features along the channels. This approach lacks a dedicated 3D feature extractor and fails to leverage the advantages of 3D information fully. Methods~\cite{horwitz2023back,cao2023complementary,wang2023multimodal} based on memory bank remain effective in 3D anomaly detection. In this regard, RGB features are extracted using pre-trained networks. Experiments show that employing manually crafted features (FPFH) for point clouds as feature extractors yields promising results. Through concatenation, the 2D and 3D features are fused. Additionally, CPMF~\cite{cao2023complementary} uses multi-view 2D features rendered as pseudo 3D features, leading to improved results. M3DM~\cite{wang2023multimodal} opts not to use manual features and instead employs point cloud features based on pre-trained networks. When addressing fusion challenges, an unsupervised training approach is adopted, impacting the relationship between RGB and point cloud features. However, it used three memory banks, significantly slowing the network speed.

The above research underscores the need for an effective approach to handle multi-modal features. In this context, we utilize depth maps as 3D features to resolve feature alignment issues. And we introduces an importance scoring module to autonomously learn the importance of the RGB feature and the depth feature. By weighting both features, we obtained the fusion result of these two features.

\begin{figure}[t]
  \centering
  \includegraphics[width=1\linewidth]{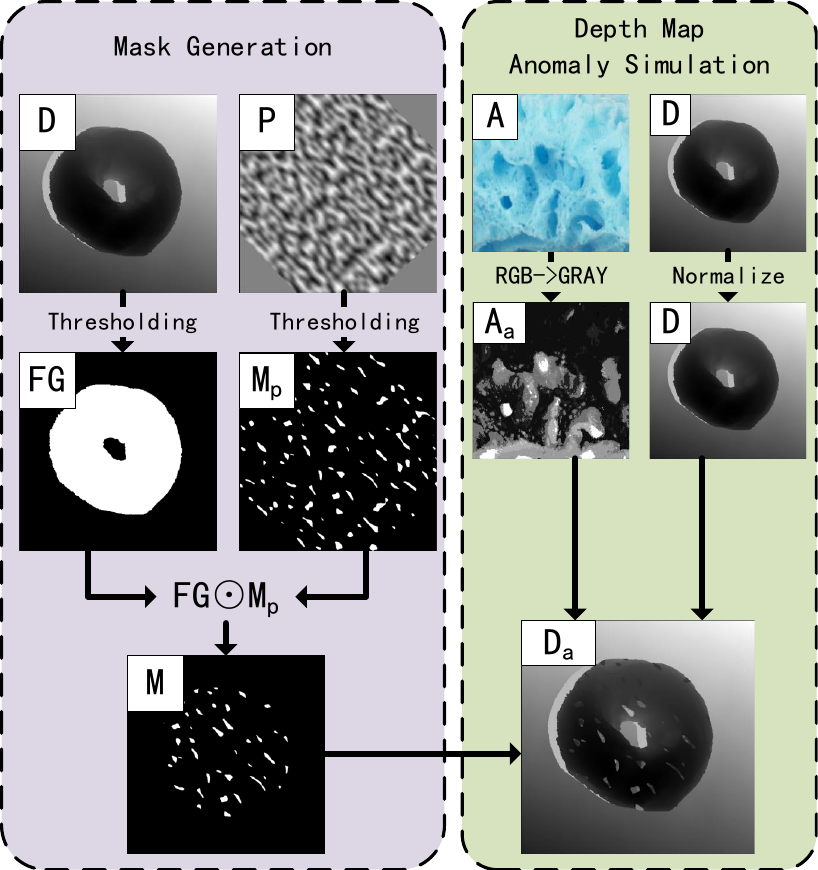}
  \caption{Depth Map Anomaly Simulation. By thresholding, the binary mask $M_p$ and $FG$ are generated from Perlin noise $P$ and depth map $D$, respectively. The final mask $M$ is obtained through the Hadamard product of the $FG$ and $M_p$. The abnormal regions are sampled from $A_a$ according to $M$ and placed on the anomaly-free image $D$ to generate the abnormal image $Da$.}
  \label{fig:2}
\end{figure}

\section{METHOD}

\subsection{Overview}

\begin{figure*}[t]
  \centering
    \includegraphics[width=1\linewidth]{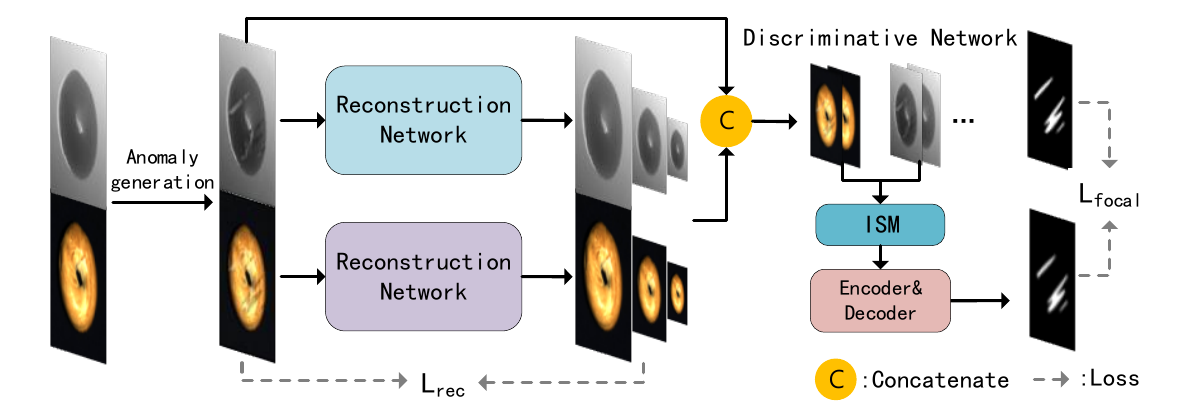}
   \caption{Overview of the proposed DBRN. DBRN receives RGB image and depth map subjected to anomaly simulation and reconstructs multi-scale normal images. Afterward, the original image and the reconstructed image are concatenated and fed into ISM to fuse these two modalities. Finally, a discriminative network is employed to obtain the anomaly segmentation result. Note that anomaly generation is not required during inference.}
   \label{fig:net}
\end{figure*}
The DBRN proposed in this paper consists of a reconstruction and discriminative networks (as shown in Figure~\ref{fig:net}). During training, abnormal samples are artificially generated based on normal images. We use a dual-branch reconstruction network to reconstruct images containing multi-scale anomaly-free content from these abnormal samples. The original and reconstructed images are concatenated and fed into the ISM for feature fusion. Finally, the discriminative network obtains segmentation results and accurately identifies defect locations by synthesizing information across multiple scales.

\subsection{Simulated anomaly generation}

We use Perlin noise to generate simulated anomaly positions (Figure~\ref{fig:2}, Mask Generation). In the experiments, we recognized that simulating defects outside of objects could introduce noise that hampers network learning.
To address this question, we only generate defects in the foreground region. We generated a foreground mask $FG$ using depth maps $D$ and combined it with the Perlin noise mask $M_p$ to create the final mask $M$ as the ground truth of the discriminative task.

For simulating defects in the depth map ( Figure ~\ref{fig:2}, Depth Map Anomaly Simulation), we performed the following steps.
By transforming image $A$ from a natural dataset, we obtain grayscale image $A_a$ to simulate possible texture features in the depth map.
To simulate protrusions and indentations outside the depth range, we normalized the depth map between 0.2 and 0.8, with values beyond this range representing protrusions and indentations.
In simulating anomalies in RGB images, we also use a random enhancement algorithm to obtain enhanced images $A_a$, but normalization is not required.

We divided anomaly simulation into three scenarios, determined by random values. 
Here we only describe the method of simulating defects in the depth map. The method for simulating defects on RGB image is similar.
The augmented image $D_a$ is produced according to the following formula: 
 \begin{equation}
 \small
 	\begin{aligned}
 		D_a = 
        \begin{cases}
           D               & \text{if } \ p\ge \varepsilon_1 \,;\\
           D\odot (1-M)    & \text{if } \ \varepsilon_2\le  p< \varepsilon_1 \,;\\
           D\odot(1-M) + \beta(D\odot M) \\+ (1-\beta)(A_{a}\odot M)  & \text{otherwise} \,. \\
        \end{cases}
 	\end{aligned}
 	\label{eq_s}
 \end{equation}
where $p$ is a random value sampling uniformly from an interval, i.e., $p \in [0,1]$, $\varepsilon_1,\varepsilon_2$ represent the probabilities for different scenarios. In our setup, $\varepsilon_1=0.7,\varepsilon_2=0.5$. $\beta$ is the opacity parameter in blending and $\beta \in [0,1]$. $\odot$ is the element-wise
multiplication operation.
\subsection{Reconstruction Network}

The reconstruction network is responsible for restoring abnormal regions to normal patterns. Given the RGB-D input, we employ a dual-branch network for separate reconstruction tasks. 
Furthermore, we use the Swin-Unet~\cite{cao2022swin} architecture as the reconstruction network framework and generate multi-scale reconstructed images. We employ multiple heads to recover the outputs from different scales of the decoder, thereby obtaining reconstruction results of various scales. It is advantageous for the discriminative network to make accurate judgments about defects of different sizes.
Generally, image similarity is evaluated using L2 and SSIM loss~\cite{wang2004image} calculations.
Due to the imbalance in the number of simulated defective samples and normal samples, we reweight the loss associated with defect positions.
Through this strategy, we enhance the ability to reconstruct abnormal regions, resulting in significant improvements in effectiveness.

The entire image is represented as a collection of pixels denoted as set $N$, which is divided into three subsets based on pixel type: $N_{bg}$ (background pixels), $N_{fg}$ (foreground pixels), and $N_{mask}$ (mask pixels).
Note that the foreground does not include the mask region. Now, the reconstruction loss is as follows:
\begin{equation}
\small
    \begin{aligned}
        L_{rec}(I,I_r) =& \frac{1}{|N|}(\lambda_{1}\sum_{x\in N_{bg}}l_2(I^x,I^x_r)+\lambda_{2}\sum_{x\in N_{fg}}l_2(I^x,I^x_r)\\
        &+\lambda_{3}\sum_{x\in N_{mask}}l_2(I^x,I^x_r))+\lambda_{4}L_{SSIM}(I,I_r)
    \end{aligned}
    \label{eq_s}
\end{equation}
where $\lambda_{1},\lambda_{2},\lambda_{3}$ and $\lambda_{4}$ are loss balancing hyper-parameters and we set $\lambda_{1}=1,\lambda_{2}=4,\lambda_{3}=10,\lambda_{4}=1$  based on the quantity of different sets. $I,I_r$ are the input image and the reconstructed image output by the network. By weighting the loss, we improved the reconstruction effect of defect locations.
\subsection{Discriminative Network}

\begin{figure}[t]
  \centering
    \includegraphics[width=1\linewidth,trim=20 0 0 0]{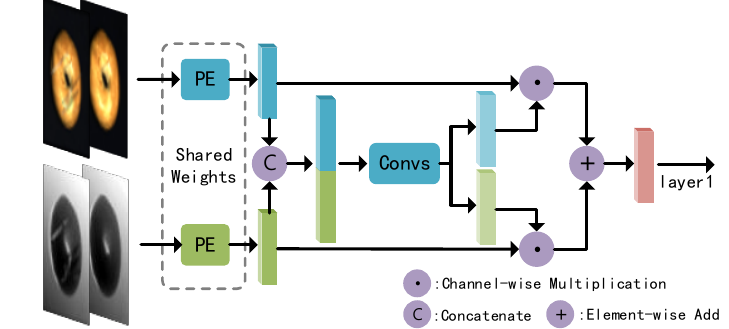}
   \caption{The importance scoring module (ISM) used by DBRN.}
   \label{fig:ISM}
\end{figure}

The role of the discriminative network is to comprehensively assess the multi-scale results of a reconstruction network for both RGB and depth modalities. 
Instead of using separate networks for each modality, we use a single discriminative network to assess the differences comprehensively. We introduce a mechanism called ISM to merge the features from both modalities (as shown in the Figure~\ref{fig:ISM}).

\subsubsection{Importance Scoring Module}

We insert ISM after the Patch Embed (PE) layer and before entering the primary layer of the discriminative network. The PE layer encodes the origin image and the reconstructed image, capturing differences between the two images in the encoded information. 
Due to the reconstruction network restoring the defective position to the normal mode, it is easy to identify the differences in the corresponding location. Therefore, the role of ISM is to discern whether the differences at the corresponding positions are significant.

The features of RGB and depth are equivalent in subsequent network layers, primarily owing to their similar distribution range and the fact that both undergo encoding through the same PE layer.  Therefore, in subsequent network layers, it is only necessary to assess whether there are differences at corresponding feature positions. Because the two modalities have different channel counts, we split the RGB channels into individual channels for processing and, finally, take the mean as the RGB embedding. 
We acquire the RGB joint feature denoted as $F_{RGB}$ and the depth joint feature as $F_D$. These two features are concatenated, and subsequently, a Convs module is applied to obtain $W_C$. $W_C$ is then divided into two parts, namely $W_{RGB}$ and $W_D$.
\begin{equation}
\small
    \begin{aligned}
        W_{RGB},W_{D}=\mathcal{F}_{split}(\mathcal{F}_{Convs}(Cat(F_{RGB},F_D)))\\
    \end{aligned}
    \label{eq_l}
\end{equation}
where $\mathcal{F}_{Convs}$ denotes the Convs module. The final features are obtained through weighting:
\begin{equation}
\small
    \begin{aligned}
        F_{fusion}=F_{RGB}\odot W_{RGB}+ F_D\odot W_{D}
    \end{aligned}
    \label{eq_l}
\end{equation}
where $\odot$ denotes channel-wise multiplication.

During training, we randomly input two identical original images concatenated in one modality, instead the concatenation of the original image and the reconstructed image.
This encourages ISM's weights to lean towards the another modality with differences. Through this approach, we achieve a selection mechanism for features from RGB and depth, propagating features from locations with pronounced differences.

\subsubsection{Multi-Scale Discrimination}

We comprehensively assess the multi-scale results of the reconstruction network to enhance the discriminative ability for defects of different sizes.
We obtain images with varying degrees of blurriness by directly enlarging the reconstructed outputs at multiple scales to the original image size. Specifically, when dealing with larger defects, relatively blurry images are more easily distinguishable, and vice versa.
Furthermore, we shared the discriminative network across multiple scales, significantly reducing computational overhead. The discriminative network employs focal loss~\cite{lin2017focal}, where we adjusted the weights for the normal and anomaly classes to mitigate the issue of sample imbalance.

In summary, DBRN accomplishes the tasks of image reconstruction and semantic discriminative. The total loss is
\begin{equation}
\small
    \begin{aligned}
        L(R,R_r,D,D_r,M,M_{out})=&L_{rec}(R,R_r)+L_{rec}(D,D_r)\\
        &+L_{focal}(M,M_{out})
    \end{aligned}
    \label{eq_l}
\end{equation}

where $R$ and $D$ are the input of RGB image and depth map. $R_r$ and $D_r$ refer to the corresponding reconstructed images.
$M$ and $M_{out}$ are the ground truth and the output anomaly segmentation masks, respectively.

\section{Experiments}

\begin{table*}[t]
\caption{Image-level AUROC (\%) score for anomaly detection of all categories of MVTec-3D AD for RGB and RGB+3D data. Best results per data domain are in bold, and the second best in underline. A * indicates that we used a reimplementation. MS indicating the use of multiple scales.}
\small

\centering

\begin{tabular}{c|ccccccccccc|c}
\hline
                         & Method    & Bagel          & \makecell[c]{Cable\\gland}  & Carrot         & Cookie         & Dowel         & Foam           & Peach           & Potato         & Rope           & Tire            & Mean             \\ \hline
\multirow{6}{*}{\rotatebox{90}{RGB}}     & PatchCore & 0.876          & 0.880          & 0.791          & 0.682          & 0.912         & 0.701          & 0.695           & 0.618          & 0.841          & 0.702           & 0.770           \\
                         & PADiM     & \underline{0.975}    & 0.775          & 0.698          & 0.582          & 0.959         & 0.663          & \underline{0.858}     & 0.535          & 0.832          & 0.760            & 0.764           \\
                         & CS-Flow   & 0.941          & \underline{0.930}     & 0.827          & 0.795          & \textbf{0.990} & 0.886          & 0.731           & 0.471          & 0.986          & 0.745           & 0.830           \\
                         & AST       & 0.947          & 0.928          & \underline{0.851}    & 0.825          & \underline{0.981}   & \textbf{0.951} & \textbf{0.895}  & 0.613          & \underline{0.992}    & \underline{0.821}     & \textbf{0.880}  \\
                         & DRAEM*    & \textbf{0.988} & \textbf{0.952} & 0.799          & \underline{0.883}    & 0.853         & \underline{0.928}    & 0.713           & \underline{0.766}    & \textbf{0.994} & \textbf{0.880}   & \underline{0.876}           \\
                         & \textbf{Ours}      & 0.936          & 0.916          & \textbf{0.867} & \textbf{0.917} & 0.866         & 0.917          & 0.795           & \textbf{0.792} & 0.942          & 0.810            & \underline{0.876}     \\ \hline
\multirow{6}{*}{\rotatebox{90}{RGB+3D}} & BTF       & 0.918          & 0.748          & 0.967          & 0.883          & 0.932         & 0.582          & 0.896           & 0.912          & 0.921          & \underline{0.886}     & 0.865           \\
                         & AST       & 0.983          & 0.873          & 0.976          & 0.971          & 0.932         & 0.885          & 0.974           & 0.981          & \textbf{1.000}     & 0.797           & 0.937           \\
                         & M3DM      & \underline{0.994}    & \textbf{0.909} & 0.972          & 0.976          & \textbf{0.960} & \underline{0.942}    & 0.973           & 0.899          & 0.972          & 0.850            & \underline{0.945}     \\
                         & CPMF      & 0.983          & \underline{0.889}   & 0.989         & 0.991          & \underline{0.958}  & 0.809         & \textbf{0.988} & 0.959          & 0.9792         & \textbf{0.969} & \textbf{0.951} \\
                         & \textbf{Ours}      & \textbf{0.998} & 0.823          & \textbf{0.998} & \textbf{0.998} & 0.821         & 0.918          & 0.973           & \underline{0.989}    & \underline{0.994}    & 0.696           & 0.921           \\
                         & \textbf{Ours(MS)}  & \textbf{0.998} & 0.836          & \underline{0.996}    & \underline{0.995}    & 0.825         & \textbf{0.959} & \underline{0.983}     & \textbf{0.993} & 0.985          & 0.710            & 0.928            \\ \hline
\end{tabular}
\label{tab:result}
\end{table*}
\subsection{Dataset}
To demonstrate the advantages of our method in 3D industrial anomaly detection, we conducted experiments on the MVTec 3D-AD dataset~\cite{bergmann2021mvtec}, the first 3D industrial anomaly detection dataset.
The dataset comprises 2,656 training samples and 1,137 testing samples. 
The training set exclusively contains defect-free data, while the testing set includes defect-free and defective samples and pixel-level labels indicating defect positions.
The MVTec 3D-AD dataset encompasses 10 categories, including scratch, hole, cut, discoloration, and others, predominantly representing geometric defects. Each category provides 2D RGB images and 3D point cloud scans. The 3D scans are acquired using structured light industrial sensors, and the resolutions of both modalities are aligned, ranging from 400 to 800 pixels. The 3D positional information is stored in a 3-channel tensor representing the x, y, and z coordinates. For our approach, the DBRN exclusively utilizes the z-channel information.
\subsection{Implementation Details}

\subsubsection{Data Preprocess.}
From the point cloud scan, we have obtained the depth component. The scan contains missing noise due to accuracy issues and shadow noise that arises from the principles of structured light imaging and cannot be captured. We generated the preprocessed depth map and foreground mask by employing methods derived from AST~\cite{rudolph2023asymmetric}. A distinction lies in how we treat the remaining zero values after inpainting: we fill them with the background mean value. This background mean value is obtained by clustering the depth map. The threshold for the foreground mask is not universally set. Instead, it is established based on the heights of different objects.
\subsubsection{Reconstruction Network}

We employ Swin-Unet~\cite{cao2022swin} architectures based on the Swin Transformer for our reconstruction tasks. Due to the robust feature extraction capabilities of the Swin-Transformer, we adjust the layer configuration of the encoder and decoder to [2, 2, 4, 2] to reduce computational load.
The same two networks are utilized for reconstructing the RGB and depth branches, respectively. We modify the input channel dimensions of PE to accommodate the distinct inputs. We apply dropout to the encoder outputs to enhance the reconstruction outcomes before feeding them into the decoder. The dropout rates for the shallow, middle, and deep layers were set at [0.6, 0.4, 0.2], respectively. During testing, these dropout layers were also activated. The learning rate for the reconstruction networks is set as $1\times 10^{-4}$.

\begin{table*}[t]
\small
\centering
\caption{Pixel-level AUROC (\%) score for anomaly detection of all categories of MVTec-3D AD for RGB and RGB+3D data. Best results per data domain are in bold, and the second best in underline. A * indicates that we used a reimplementation. MS indicating the use of multiple scales.}
\begin{tabular}{c|ccccccccccc|c}
\hline
                         & Method    & Bagel          & \makecell[c]{Cable\\gland}  & Carrot         & Cookie         & Dowel         & Foam           & Peach           & Potato         & Rope           & Tire            & Mean             \\ \hline
\multirow{3}{*}{\rotatebox{90}{RGB}}  & PatchCore & 0.983          & \textbf{0.984} & \underline{0.98}     & \textbf{0.974} & \textbf{0.985} & 0.836          & \underline{0.976}    & \underline{0.982}    & \underline{0.989}    & \textbf{0.975} & \underline{0.9664}    \\
                      & DRAEM*    & \textbf{0.992} & 0.851          & 0.964          & \underline{0.949}    & 0.976          & \textbf{0.956} & 0.934          & 0.977          & \textbf{0.996} & 0.942          & 0.9537          \\
                      & \textbf{Ours}      & \underline{0.985}    & \underline{0.952}    & \textbf{0.991} & 0.939          & \underline{0.98}     & \underline{0.946}    & \textbf{0.979} & \textbf{0.99}  & 0.986          & \underline{0.973}    & \textbf{0.9721} \\ \hline
\multirow{5}{*}{\rotatebox{90}{RGB+3D}} & BTF       & \textbf{0.996} & \underline{0.992}    & \underline{0.997}    & \textbf{0.994} & \underline{0.981}    & \underline{0.974}    & \underline{0.996}    & \underline{0.998}    & 0.994          & \underline{0.995}    & \underline{0.9917}    \\
                      & AST       & 0.988          & 0.984          & \underline{0.997}    & 0.957          & 0.931          & 0.971          & \textbf{0.997} & \underline{0.998}    & 0.941          & 0.988          & 0.9752          \\
                      & M3DM      & \underline{0.995}    & \textbf{0.993} & \underline{0.997}    & 0.985          & \textbf{0.985} & \textbf{0.984} & \underline{0.996}    & 0.994          & \textbf{0.997} & \textbf{0.996} & \textbf{0.9922} \\
                      & Ours      & \textbf{0.996} & 0.937          & \textbf{0.998} & 0.985          & 0.974          & 0.96           & 0.995          & \textbf{0.999} & \underline{0.996}    & 0.964          & 0.9804          \\
                      & \textbf{Ours(MS)} & 0.993          & 0.932          & \underline{0.997}    & \underline{0.99}     & 0.966          & \textbf{0.984} & \textbf{0.997} & \textbf{0.999} & 0.992          & 0.976          & 0.9826          \\ \hline
\end{tabular}

\label{tab:result2}
\end{table*}
\subsubsection{Discriminative Network}

The discriminator networks utilize UNet~\cite{ronneberger2015u} architectures with skip connections.
Multiple scale results from the reconstruction are resized to a uniform size concatenated along the batch dimension to enable weight sharing. The final result is obtained by averaging the multi-scale outcomes. The ISM is positioned between the PE layer and layer1. The Convs module involves a 2-layer Conv2D operation with batch normalization and an activation layer. The convolutional kernel size is set to 3, with a stride of 1 and padding of 1. The hidden layer's dimension matches the input feature dimension. For the discriminator network, the learning rate is set as $1\times 10^{-5}$. To expedite the ISM module's training speed, the convolutional layers' learning rate is specifically set as $1\times 10^{-3}$.

\subsubsection{Evaluation Metrics.}
For general anomaly detection methods, we evaluate the image-level anomaly detection performance with the area under the receiver operator curve. For simplicity, Image-level AUROC is denoted as I-AUROC, and higher I-AUROC means better image-level anomaly detection performance. Like I-AUROC,  the receiver operator curve of pixel-level predictions can be used to compute the Pixel-level AUROC (P-AUROC) to evaluate the segmentation performance.

\subsection{Anomaly Detection on MVTec 3D-AD.}
Table~\ref{tab:result} presents the I-AUROC results of various methods grouped by RGB and RGB+3D modalities on the MVTEC 3D-AD dataset. Table~\ref{tab:result2} displays the segmentation results using P-AUROC.

For RGB anomaly detection, we compare several methods that perform well in 2D anomaly detection, most of which employ a large pre-trained model. Additionally, we replicate the DRAEM~\cite{zavrtanik2021draem} method on the MVTEC 3D-AD dataset. While its performance is suboptimal on the MVTEC-AD~\cite{bergmann2019mvtec} dataset, it proves more effective than Patchcore~\cite{roth2022towards}, Padim~\cite{defard2021padim}, and CS-Flow~\cite{rudolph2022fully} on the 3D dataset. Our approach can also utilize only RGB images. Compared to DRAEM, we enhance anomaly simulation and employ a transformer-based reconstruction network. Although there is no improvement in scores for image-level classification, there is a 1.8\% increase in anomaly localization accuracy.

When using both RGB and 3D information, DBRN achieved state-of-the-art performance without a large pre-trained model and a memory bank on the MVTEC 3D-AD dataset. Specifically, we outperformed the BTF method by 6.3\% in I-AUROC and achieved competitive results in most categories. BTF~\cite{horwitz2023back}, AST~\cite{rudolph2023asymmetric}, M3DM~\cite{wang2023multimodal}, and CPMF~\cite{cao2023complementary} employed large pre-trained models or memory libraries. Among these, M3DM~\cite{wang2023multimodal} used 3 memory banks, consuming substantial memory. Our speed comparison with these methods can be found in Section 4.5. Compared to using only RGB, our approach yielded a 5.2\% improvement, demonstrating the positive impact of incorporating 3D information.

\begin{figure}[t]
  \centering
    \includegraphics[width=1\linewidth,trim=20 0 20 0]{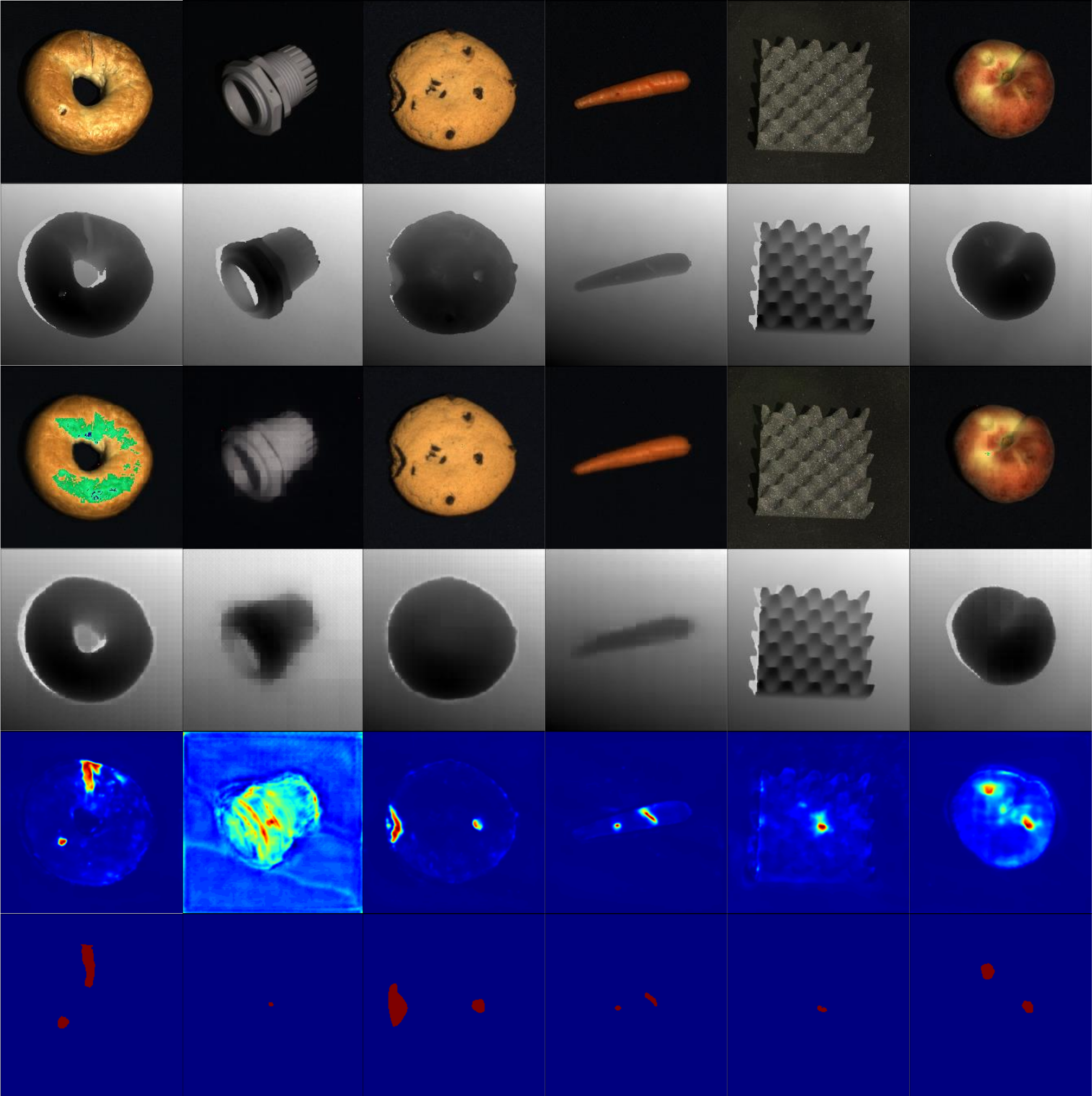}
   \caption{Qualitative examples. The first and second rows are the input RGB image and the depth map. The third and fourth rows are reconstruction results. The fifth and sixth rows are the prediction results and ground truth, respectively.}
   \label{fig:result}
\end{figure}

As depicted in the Figure~\ref{fig:result}, DBRN's reconstruction results and discriminative outcomes are displayed for various categories. When faced with object defects, the reconstruction network can fill in corresponding positions, aiding in defect localization, as in the case of cookies. This yields superior performance compared to other methods. The results indicate that DBRN's reconstruction scheme excels with objects featuring similar textures and simple structures. However, when confronted with complex structures, the anomaly simulation fails to yield common defect patterns effectively. 
Consequently, the reconstruction network cannot reconstruct the defect's location, leading to an inability of the discriminative network to differentiate anomalies.
\subsection{Ablation Study.}

\begin{table}[t]
\centering
\caption{Image/Pixel-level AUROC (\%) of the proposed method with various reconstruction network and discriminative network on MVTec 3D-AD dataset.}

\begin{tabular}{ccc}
\hline
Net                                 & I-AUROC & P-AUROC \\ \hline
DBRN( Res-UNet + Swin-Unet ) & 0.843 & 0.958 \\
DBRN( Res-UNet + UNet )      & 0.911 & 0.976 \\
DBRN( Swin-Unet + Swin-Unet )     & 0.875 & 0.965 \\
DBRN( Swin-Unet + UNet )           & \textbf{0.921} & \textbf{0.980}
 \\ \hline
\end{tabular}

\label{tab:net_ab}
\end{table}

\textbf{Architecture.}
Table~\ref{tab:net_ab} presents the anomaly detection performance of different architectural configurations for the reconstruction and discriminative networks. The reconstruction network employs a dual-branch structure, wherein we have choosen the Swin-Unet architecture based on the transformer model and the representative Res-UNet~\cite{8589312} architecture based on CNN for comparison. 
The discriminator network is configured with a single-branch setup, utilizing the UNet architecture, and it is compared with the Swin-Unet architecture.
The transformer architecture can capture global contextual information, whereas CNNs focus on local details.
The results indicate that the architecture based on Transformer is more suitable for the reconstruction network, achieving a 1\% higher performance compared to the CNN architecture.
As in Figure~\ref{fig:result}, the model based on the transformer can effectively restore phenomena like partial object loss, which is challenging for CNNs to address. 
For the discriminative network, what we need is to determine whether local texture features are different. Shallow CNNs tend to perform better in this context, which is 4.6\% higher than the use of Swin-Unet.
\begin{table}[]
\centering
\caption{Image/Pixel-level AUROC (\%) of the proposed method according to $L_{fg}$ and $L_{mask}$ on MVTec 3D-AD dataset.}
\begin{tabular}{cccc}
\hline
$L_{fg}$  & $L_{mask}$ & I-AUROC & P-AUROC \\ \hline
        &          & 0.912 & 0.969 \\
$\surd$ &          & 0.920 & 0.970 \\
        & $\surd$  & 0.917 & 0.977 \\
$\surd$ & $\surd$  & \textbf{0.921} & \textbf{0.980} \\ \hline
\end{tabular}

\label{tab:loss_ab}
\end{table}

\textbf{Loss function.}
Table~\ref{tab:loss_ab} demonstrates the impact of loss enhancement strategies. In regression tasks, sample imbalance can also have adverse effects. We achieve better results by adjusting the quantity balance between positive and negative samples. We focus on reconstructing objects rather than the background for the reconstruction task. By assigning a higher weight to the foreground ($L_{fg}$), we improve the I-AUROC score by 0.8\%. However, our primary goal was to train the network to distinguish between positive and negative samples. To address the extreme imbalance of these samples, we introduce $L_{mask}$. Adding this loss effectively addressed the imbalance issue, resulting in a 0.8\% performance improvement. Ultimately, we achieve the best performance by simultaneously utilizing both loss functions.

\begin{table}[t]
\centering
\caption{Ablation study on fusion module. The first row is the Early-Fusion. The second and third rows are the Deep-Fusion, representing element-wise multiplication (Hadamard product) and concatenation, respectively.}
\begin{tabular}{ccc}
\hline
 Method                 & I-AUROC & P-AUROC \\ \hline
Cat$(rgbd, rgbd) $      & 0.882 & 0.966 \\
$f(rgb)\odot f(d)$      & 0.891 & 0.976 \\
Cat$(f(rgb),f(d))$      & 0.898 & 0.974 \\
\textbf{ISM}                     & \textbf{0.921} & \textbf{0.980} \\ \hline
\end{tabular}

\label{tab:fusion_ab}
\end{table}

\textbf{Fusion method.}
Table~\ref{tab:fusion_ab} demonstrates the impact of the loss strategy on performance. Based on~\cite{huang2022multi}, we adopt Early-Fusion and Deep-Fusion strategies for modal fusion. For Early-Fusion, we directly concatenate the original images and reconstructed images in sequence, allowing the network to learn the required features. However, this method demands a high learning capacity from the network and thus yields poor performance.
In the Deep-Fusion approach, we explore two methods: element-wise Hadamard product and concatenation (Cat). After the cat operation, we use linear layers to restore the feature dimension to its original size. These two fusion methods outperform Early-Fusion at the feature level but still rely on the network's learning capability.
Our method also falls under the Deep-Fusion category. In this phase, we incorporate features into the discriminative network before entering the encoder using random perturbation input and a scoring mechanism. This facilitates the network's learning process and results in the best performance, which is 2.3\% higher than the second
best one.

\subsection{Inference Time.}

Table~\ref{tab:fps_ab} displays a comparison of speed and parameter quantities for our several architecture networks, as well as AST~\cite{rudolph2023asymmetric}, BTF~\cite{horwitz2023back} and M3DM~\cite{wang2023multimodal}. The machine's hardware configuration used for inference is NVIDIA GeForce RTX 3060. Due to its more extensive scale, the ResNet architecture achieves only about 6  fps and does not yield excellent results. When using the UNet as the discriminative network, the Swin architecture shows a 1\% performance improvement, significantly reducing parameters and computational complexity. This indicates that the transformer structure is superior in reconstruction tasks. We calculated AST's parameter quantity and computational complexity and found that our method achieves a 170\% speed improvement with results that are only 1.6\% lower in score than AST. However, BTF and M3DM achieve only 1.43 fps and 0.24 fps, respectively, because of large models and memory banks, making it unacceptable for practical industrial use. In contrast, our model performs excellently in speed and maintains high accuracy, making it suitable for deployment.
\begin{table}[t]
\centering
\caption{Inference Time.}
\begin{tabular}{ccccc}
\hline
Methods                       & I-AUROC  & FPS            \\\hline
DBRN( Res-UNet + Swin-Unet )  & 0.843  & 6.91           \\
DBRN( Res-UNet + UNet )       & 0.911  & 6.64           \\
DBRN( Swin-Unet + Swin-Unet )     & 0.875  & 37.46          \\
\textbf{DBRN( Swin-Unet + UNet )}          & \textbf{0.921}  & \textbf{48.93} \\
AST                           & 0.9372 & 10.39          \\
BTF                         & 0.8645 & 1.49        \\
M3DM                         & 0.9447 & 0.24        \\\hline  
\end{tabular}

\label{tab:fps_ab}
\end{table}

\section{Conclusion}

In this paper, we introduce a 3D anomaly detection model called DBRN based on RGB-D data. This method relies on a reconstruction approach, utilizing a dual-branch network to reconstruct RGB images and depth maps separately. We abandon point clouds to use depth data, which retains positional information and allows 2D methods to process depth information, eliminating alignment issues between RGB and depth data. Additionally, we propose the ISM feature fusion module to provide effective differentiation information for the discriminative network. Underpinned by a transformer-based reconstruction network and a U-Net-based discriminative network, DBRN achieves an AUROC of 92.8\% on the MVTEC 3D-AD dataset and exhibits excellent performance across multiple classes. Furthermore, we achieve the fastest inference speed without using large pre-trained models and memory banks.

One of the shortcomings of DBRN is its inability to reconstruct defects when dealing with structurally complex objects. In our future work, we will focus on enhancing the performance of the reconstruction network and refining anomaly simulation strategies.

{\small
\bibliographystyle{ieee_fullname}
\bibliography{egbib}
}

\end{document}